\documentclass[10pt,twocolumn,letterpaper]{article}
\usepackage{float}
\usepackage{comment}
\usepackage[pagenumbers]{cvpr}
\newcommand{\red}[1]{{\color{red}#1}}

\usepackage{color}
\usepackage{xcolor}
\definecolor{deepGreen}{RGB}{0,153,0}
\definecolor{orange}{RGB}{255,125,0}
\def\red#1{\textcolor[rgb]{1,0,0}{#1}}

\def\green#1{\textcolor[rgb]{0,1,0}{#1}}

\definecolor{sainone}{RGB}{236, 242, 249} 
\definecolor{saintwo}{RGB}{255, 230, 204}

\newcommand{\cut}[1]{}
\usepackage{multirow}
\usepackage{amsmath}
\usepackage{amssymb}
\usepackage{mathtools}
\usepackage{arydshln}
\usepackage{soul}
\usepackage{nicefrac}
\usepackage{booktabs}
\usepackage{stmaryrd}
\usepackage{caption}
\usepackage{graphicx}
\usepackage{colortbl}
\definecolor{gray}{gray}{0.9}
\definecolor{pink}{RGB}{255, 234, 232}
\sethlcolor{pink}
\usepackage[accsupp]{axessibility}
\usepackage{scalerel,graphicx,xparse}

\makeatletter
\apptocmd\@maketitle{{\myfigure{}\par}}{}{}
\makeatother
\usepackage{capt-of,etoolbox}

\makeatletter
\newcommand\notsotiny{\@setfontsize\notsotiny\@vipt\@viipt}
\makeatother

\usepackage{xcolor,pifont}
\newcommand*\colourcheck[1]{%
  \expandafter\newcommand\csname #1check\endcsname{\textcolor{#1}{\ding{52}}}%
}
\newcommand*\colourcross[1]{%
  \expandafter\newcommand\csname #1cross\endcsname{\textcolor{#1}{\ding{55}}}%
}
\colourcheck{blue}
\colourcheck{green}
\colourcheck{black}
\colourcross{red}
\colourcross{black}
\usepackage{anyfontsize}
\usepackage{fontawesome5}
\usepackage{wrapfig}

\definecolor{cvprblue}{rgb}{0.21,0.49,0.74}
\usepackage[pagebackref,breaklinks,colorlinks,allcolors=cvprblue]{hyperref}

\title{\vspace{-1cm}SketchFusion: Learning Universal Sketch Features through \\ Fusing Foundation Models\vspace{-8mm}}

\author{Subhadeep Koley\textsuperscript{1,2}\thanks{Equal contribution} \hspace{.2cm} Tapas Kumar Dutta\textsuperscript{1}$^*$ \hspace{.2cm} Aneeshan Sain\textsuperscript{1} \hspace{.2cm}  Pinaki Nath Chowdhury\textsuperscript{1}\\ Ayan Kumar Bhunia\textsuperscript{1}\hspace{.3cm}  Yi-Zhe Song\textsuperscript{1,2} \\
\textsuperscript{1}SketchX, CVSSP, University of Surrey, United Kingdom.  \\
\textsuperscript{2}iFlyTek-Surrey Joint Research Centre on Artificial Intelligence.\\
{\tt\small \{s.koley, a.sain, p.chowdhury, a.bhunia, y.song\}@surrey.ac.uk}
}

\newcommand\myfigure{
\centering
\vspace{-0.9cm}
\captionsetup{type=figure}
    \includegraphics[width=1\textwidth]{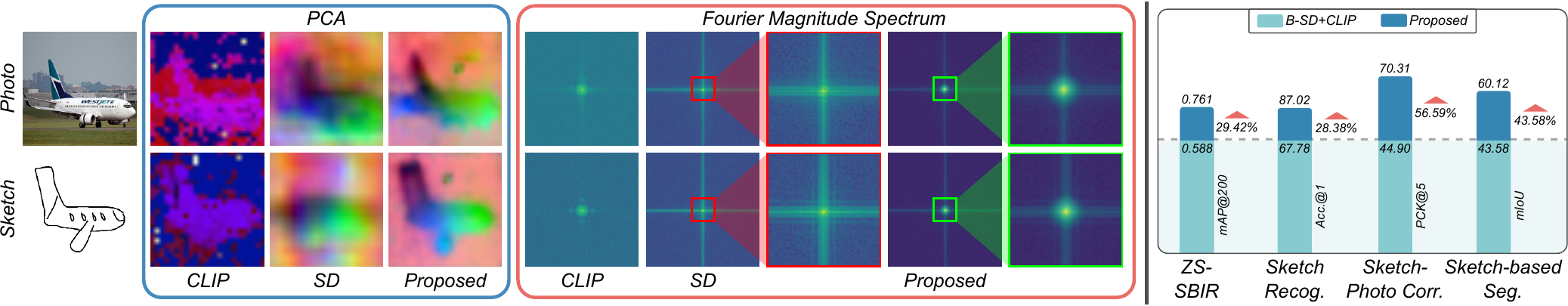}
    \vspace{-0.6cm}
    \captionof{figure}{\textit{(Left):} Apart from high-resolution image generation, text-to-image diffusion models (\eg, Stable diffusion (SD) \cite{rombach2022high}) with their innate object understanding capability \cite{zhang2024tale, tang2023emergent}, have shown remarkable performance across a wide range of \textit{image-based} vision tasks (\eg, segmentation \cite{xu2023open}, depth estimation \cite{zhao2023unleashing}, etc.). However, upon analysing the PCA representation of SD's intermediate UNet features, we observe that it struggles to achieve similar results when working with freehand \textit{abstract} sketches (detail in \cref{sec:pilot}). Unlike \textit{pixel-perfect} photos, \textit{highly abstract} freehand sketches are sparse and lack detailed textures and colours \cite{eitz2012humans}, making it harder for the SD model to extract meaningful features. Furthermore, investigating the SD denoising process in the \textit{frequency} domain (via Fourier Transform), we observe the predominance of high-frequency (HF) components, rather than their low-frequency (LF) counterpart -- crucial for capturing comprehensive semantic context. To mitigate this inherent bias within SD, we reinforce the diffusion process with another pretrained model (\ie, CLIP \cite{radford2021learning}) whose bias is complementary (\ie, focuses on LF) to SD. Consequently, the proposed extractor can extract semantically meaningful and accurate features from both sketches and photos, that encapsulate a broader frequency spectrum (\ie, HF and LF). \textit{(Right:)} Testing the proposed method with different sketch-based \textit{discriminative} and \textit{dense prediction} tasks (requiring knowledge of \textit{both} sketch and image), we find a marked improvement over baseline SD+CLIP hybrid feature extractor.
    }

\label{fig:pilot}
\vspace{+0.2cm}
}
\begin{document}
\maketitle

\vspace{-0.2cm}
\begin{abstract}
While foundation models have revolutionised computer vision, their effectiveness for sketch understanding remains limited by the unique challenges of abstract, sparse visual inputs. Through systematic analysis, we uncover two fundamental limitations: Stable Diffusion (SD) struggles to extract meaningful features from abstract sketches (unlike its success with photos), and exhibits a pronounced frequency-domain bias that suppresses essential low-frequency components needed for sketch understanding. Rather than costly retraining, we address these limitations by strategically combining SD with CLIP, whose strong semantic understanding naturally compensates for SD's spatial-frequency biases. By dynamically injecting CLIP features into SD's denoising process and adaptively aggregating features across semantic levels, our method achieves state-of-the-art performance in sketch retrieval (+3.35\%), recognition (+1.06\%), segmentation (+29.42\%), and correspondence learning (+21.22\%), demonstrating the first truly universal sketch feature representation in the era of foundation models.

\vspace{-0.1cm}
\end{abstract}
\vspace{-0.3cm}

\section{Introduction}
\label{sec:intro}
\vspace{-0.2cm}

The quest for universal sketch-specific features has been a foundational challenge in computer vision \cite{xu2020deep, sangkloy2016the, sain2023clip, koley2024text}. Sketches, with their inherent characteristics of abstraction, sparsity, and cross-modal interpretation, demand fundamentally different feature representations compared to natural images \cite{yang2021sketchaa, koley2024how}. While foundation models \cite{radford2021learning, rombach2022high} have revolutionised various visual tasks, their effectiveness for sketch understanding remains largely unexplored \cite{koley2023its}, particularly given sketches' unique position at the intersection of human abstraction and computer vision \cite{eitz2012humans, koley2024how}.

Through systematic analysis, we find that even powerful foundation models like Stable Diffusion (SD) \cite{rombach2022high} face significant challenges with sketch data. Our pilot studies (\cref{sec:pilot}) reveal two fundamental limitations. First, while SD excels at extracting task-specific representative features from pixel-perfect RGB images, it struggles with abstract freehand sketches, likely due to its large-scale pretraining being predominantly focused on photographic data \cite{rombach2022high}. Second, and more intriguingly, frequency-domain analysis of SD's UNet reveals an inherent architectural bias: the model systematically promotes high-frequency (HF) components while suppressing low-frequency (LF) information -- a characteristic that proves particularly problematic for sketch understanding tasks, especially dense prediction problems like segmentation and correspondence learning.

These findings present a significant challenge. Retraining SD on sketch data risks catastrophic forgetting of its valuable pretrained knowledge \cite{koley2023its}, while developing sketch-specific architectures would fail to leverage the rich visual understanding already present in foundation models. Our key insight comes from analysing the complementary nature of different foundation models: while SD's features are spatially-aware but lack semantic accuracy, CLIP's \cite{radford2021learning} visual encoder exhibits strong semantic understanding even at the cost of spatial precision. This complementarity suggests that strategic combination \cite{zhang2024tale} of these models could yield features that capture both the detailed structure and semantic meaning necessary for sketch understanding.

Based on this insight, we propose a novel framework that leverages the complementary strengths of these foundation models. Our method systematically injects CLIP's visual features into multiple layers of SD's UNet during the denoising process, using lightweight 1D convolutions to influence feature extraction at various semantic levels. PCA analysis of the resulting features demonstrates that this combination successfully captures both the HF spatial details from SD \cite{rombach2022high} and the LF semantic components from CLIP \cite{radford2021learning}, producing rich, semantically-aware representations suited for diverse sketch-based tasks.

Moreover, we observe that different layers of SD's UNet capture features at varying levels of semantic-granularity \cite{koley2024text}, with different combinations better suited to specific downstream tasks \cite{tang2023emergent}. Rather than manually selecting optimal features, we train a dynamic feature aggregator that automatically determines the most effective combination for any given task, eliminating the need for task-specific tuning. This adaptive approach ensures optimal performance across diverse sketch-based applications without requiring manual layer selection or task-specific optimisation.

Our extensive experiments demonstrate the effectiveness of this approach across a comprehensive suite of sketch-based tasks: \textit{(i)} Sketch Recognition: $+1.06\%$ improvement over state-of-the-art, \textit{(ii)} Sketch-based Image Retrieval: $+3.35\%$ improvement in retrieval accuracy, \textit{(iii)} Sketch-based Segmentation: $+29.42\%$ boost in mean IoU, \textit{(iv)} Sketch-Photo Correspondence: $+21.22\%$ increase in matching precision. These significant improvements across all major sketch-specific tasks suggest that our method has uncovered fundamental principles for extracting effective sketch features from foundation models.

In summary, our contributions are: \textit{(i)} A systematic analysis revealing the limitations of pretrained foundation models for sketch understanding, supported by both empirical results and frequency-domain analysis, \textit{(ii)} A novel method for combining complementary foundation models that address these limitations without retraining, \textit{(iii)} An adaptive feature aggregation approach that automatically optimises representations for different downstream tasks, and \textit{(iv)} Comprehensive experimental validation demonstrating substantial improvements across all major sketch-specific tasks, establishing a new paradigm for universal sketch feature learning.

\vspace{-0.2cm}
\section{Related Works}
\label{sec:related}
\vspace{-0.2cm}
\noindent\textbf{Diffusion Model for Representation Learning.} Diffusion models \cite{ho2020denoising} have now became the de-facto standard for high-resolution 2D \cite{rombach2022high, dhariwal2021diffusion} and 3D \cite{shim2023diffusion, chou2023diffusion} image and video \cite{chen2024videocrafter2} generation. Apart from generation, it gave rise to various image editing frameworks like Prompt-to-Prompt \cite{hertz2022prompt}, Imagic \cite{kawar2023imagic}, Dreambooth \cite{ruiz2023dreambooth}, etc. The success of diffusion models across various generative tasks and their inherent {zero-shot generalisability} has led to their adoption as {efficient feature extractors} in different forms of \textit{discriminative} and \textit{dense prediction} tasks \cite{wang2024diffusion} like image retrieval \cite{koley2024text}, classification \cite{li2023your}, semantic \cite{baranchuk2021label} and panoptic \cite{xu2023open} segmentation, depth estimation \cite{zhao2023unleashing}, object detection \cite{xu20233difftection}, image-to-image translation \cite{tumanyan2023plug, koley2023its}, controllable-generation \cite{zhang2023adding, mou2023t2i} and medical imaging \cite{de2023medical}, to name a few. A few recent works have developed specialised diffusion models together with carefully crafted task-specific designs like test-time adaptation \cite{prabhudesai2023diffusion}, masked-reconstruction \cite{pan2023masked}, or compact latent generation \cite{hudson2024soda}  to improve their representation ability. Nonetheless, our pilot study (\cref{sec:pilot}) reveals that SD, often struggles to extract representative features (\cref{fig:pilot} left) from binary and sparse sketches, hampering downstream task performance. In this work, we leverage a pretrained CLIP \cite{radford2021learning} visual encoder to enhance SD's feature representation abilities for sketch-based discriminative and dense prediction tasks.

\noindent\textbf{Sketch-based Vision Applications.} \textit{``Freehand sketch"} has now established itself as a worthy interactive alternative modality to text for downstream vision tasks \cite{yelamarthi2018zero, sain2023clip, koley2023picture, chen2023democaricature, chowdhury2023what, ham2022cogs, voynov2022sketch}. With its success in sketch-based image retrieval (SBIR) \cite{sain2023clip, chowdhury2022partially, sangkloy2022sketch, dey2019doodle}, sketches are now being used in a wide array of discriminative \cite{dey2019doodle, collomosse2019query, chowdhury2023what}, generative \cite{zhang2023adding, ghosh2019interactive, koley2023picture} and dense prediction \cite{bhunia2023sketch2saliency, hu2020sketch} tasks like controllable $2D$ \cite{koley2023its, zhang2023adding}, $3D$ \cite{wangsketch2vox, bandyopadhyay2024doodle} image generation and editing \cite{koley2023picture, richardson2021encoding, ham2022cogs, zhang2023adding, mou2023t2i, mikaeili2023sked}, video synthesis \cite{xing2024tooncrafter}, garment designing \cite{li2018foldsketch, chowdhury2022garment}, segmentation \cite{hu2020sketch}, creative content generation \cite{chen2023democaricature}, object detection \cite{chowdhury2023what}, salient region prediction \cite{bhunia2023sketch2saliency}, augmented and virtual reality applications \cite{luo2022structure, luo20233d}, medical imaging \cite{kobayashi2023medical}, class incremental learning \cite{bhunia2022doodle}, or composed retrieval \cite{koley2023you}. In this paper, we aim to tackle some of the sketch-based downstream tasks by reinforcing the diffusion feature extractor via a pretrained CLIP \cite{radford2021learning} visual encoder.

\noindent\textbf{Backbones for Sketch-based Vision Applications.} Backbones for sketch-based vision applications can be broadly categorised as -- \textit{(i)} pretrained CNN or Vision Transformer (ViT) \cite{chowdhury2022partially, bhunia2022sketching, ribeiro2020sketchformer}, \textit{(ii)} Self-supervised pretraining \cite{pang2020solving, li2023photo, bhunia2021vectorization}, and \textit{(iii)} pretrained foundation models \cite{radford2021learning, rombach2022high, karras2019style}. Starting with ImageNet pretrained CNNs \cite{chowdhury2022partially, bhunia2022sketching} or JFT pretrained ViTs \cite{ribeiro2020sketchformer, sain2023exploiting} with carefully-crafted design choices (\eg, cross \cite{lin2023zero} and spatial attention \cite{song2017deep}, reinforcement learning \cite{bhunia2022sketching}, test-time training \cite{sain2022sketch3t}, meta-learning \cite{sain2021stylemeup}, knowledge distillation \cite{sain2023exploiting}, etc.), the community moved towards self-supervised pretraining either via solving pretext tasks like sketch vectorisation \cite{bhunia2021vectorization}, rasterisation \cite{bhunia2021vectorization}, solving sketch-photo jigsaw \cite{pang2020solving} or by large-scale photo pretraining \cite{li2023photo}. With the advent of large-scale foundation models, several works \cite{koley2023picture, sain2023clip, sangkloy2022sketch, koley2024text} used them as a backbone for different sketch-based tasks like image generation \cite{zhang2023adding, koley2023picture}, shape synthesis \cite{bandyopadhyay2024doodle}, or retrieval \cite{koley2024text, sangkloy2022sketch}. Foundation models have now become the gold standard for sketch-based tasks due to their strong \textit{zero-shot} performance and extensive \textit{pretrained knowledge} \cite{koley2024text, sangkloy2022sketch, koley2024how}. Among these, generative models like SD are spatially aware \cite{zhang2024tale, koley2024text}, yet less accurate for dense prediction tasks. In contrast, vision-language models like CLIP \cite{radford2021learning} offer accurate features but are inherently sparse \cite{zhang2024tale}. In this paper, we aim to get the \textit{best of both worlds} by enhancing the SD feature extraction with CLIP.

\noindent\textbf{Reinforcing Feature Extractors.} There are numerous ways of reinforcing an \textit{off-the-shelf} pretrained feature extractor to adapt them for different downstream tasks. It can broadly be categorised as -- \textit{(i)} Mixture of Experts (MoE) \cite{yu2024boosting, di2024boost}, \textit{(ii)} Prompt learning \cite{zhou2022learning, jia2022visual}, \textit{(iii)} Hybrid models \cite{xia2024vit, zhang2024tale}, \textit{(iv)} Finetuning \cite{yang2024using, sain2023clip}, and \textit{(v)} Global-local feature merging \cite{haji2024elasticdiffusion, yu2023zero}. MoE combines multiple specialised sub-models (\ie, ``experts''), each responsible for specific tasks or data patterns, to boost overall performance and adaptability \cite{di2024boost}. In contrast, prompt learning deals with learning additional trainable components in the textual \cite{zhou2022learning, zhou2022conditional} or pixel \cite{bahng2022exploring, jia2022visual} domain for task-specific adaptation. Hybrid models \cite{wang2024diffusion} combine two different kinds of models like CNN and ViT \cite{xia2024vit}, or SD and DINO \cite{zhang2024tale} to extract the best of both worlds. Another avenue deals with finetuning large-scale pretrained models with selective handcrafted data for specific tasks \cite{sain2023clip, chowdhury2023what}. Finally, a few other works \cite{haji2024elasticdiffusion, yu2023zero, wu2023clipself} amalgamate global and local features from the same \cite{wu2023clipself} or different \cite{haji2024elasticdiffusion, yu2023zero} models to improve downstream task performance. In this paper, we take the hybrid model approach and complement \textit{CNN-based} SD features with \textit{transformer-based} CLIP visual encoder features for different kinds of sketch-based vision tasks.

\vspace{-0.2cm}
\section{Preliminaries}
\vspace{-0.1cm}
\subsection{Background: CLIP}
Contrastive language-image pretraining (CLIP) \cite{radford2021learning} architecture consists of a visual and a textual encoder, trained on $\sim$$400$M image-text pairs, with the goal of learning a shared embedding manifold. Using N-pair contrastive loss, the model aligns matching image-text pairs by maximising their cosine similarity and minimises it for random unpaired ones \cite{radford2021learning}. The image encoder, typically a ViT \cite{dosovitskiy2021image}, encodes an image $\mathbf{i}$ as a feature vector $\mathbf{V}(\mathbf{i}) \in \mathbb{R}^d$. Meanwhile, the textual encoder processes a word sequence $\mathcal{W} = \{w^0, w^1, \dots, w^k\}$ through byte pair encoding \cite{radford2021learning} and a learnable word embedding layer (vocabulary size of $49, 152$), converting each word $w^i$ into a token embedding, thus creating an embedding sequence $\mathcal{W}_e = \{\mathbf{w}^0_e, \mathbf{w}^1_e, \dots, \mathbf{w}^k_e\}$ \cite{radford2021learning}. $\mathcal{W}_e$ is then passed through a transformer network $\mathbf{T}(\cdot)$, producing the final text feature $\mathbf{T}(\mathcal{W}_e) \in \mathbb{R}^d$ \cite{radford2021learning}. Once trained, the CLIP model can be used in downstream tasks to enable zero-shot capability and cross-modal understanding by aligning visual and textual data in a shared embedding space \cite{radford2021learning}.

\vspace{-0.2cm}
\subsection{Background: Stable Diffusion (SD)}
\vspace{-0.2cm}
\noindent\textbf{Overview.} The SD model \cite{rombach2022high} generates images by gradually removing noise from an initial 2D isotropic Gaussian noise image through iterative refinement. It involves two reciprocal processes -- ``forward'' and ``reverse'' diffusion \cite{rombach2022high, ho2020denoising}. The \textit{forward} process iteratively adds random Gaussian noise into a clean image $\mathbf{i}_0 \in \mathbb{R}^{h\times w\times 3}$ for $t$ timesteps to generate a noisy image $\mathbf{i}_t$ as $\mathbf{i}_t = \sqrt{\bar{\alpha}_t}\mathbf{i}_{0} + \sqrt{1-\bar{\alpha}_t}\epsilon$. Here, $\epsilon$$\sim$$\mathcal{N}(0,\mathbf{I})$, and $\{\alpha_t\}_1^T$ is a pre-defined noise schedule where $\bar{\alpha}_t=\prod_{k=1}^t \alpha_k$ with $t$$\sim$$U(0,T)$. The \textit{reverse} process involves training a denoising UNet \cite{ronneberger2015u} $\mathbf{U}_\theta$ (with an $l_2$ objective), which estimates the input noise $\epsilon \approx \mathbf{U}_\theta(\mathbf{i}_t,t)$ from the noisy image $\mathbf{i}_t$ at each $t$. After training, $\mathbf{U}_\theta$ is capable of reconstructing a clean image from a noisy input \cite{rombach2022high}. The inference process begins with a random 2D noise sample, $\mathbf{i}_T$$\sim$$\mathcal{N}(0, \mathbf{I})$. The trained model $\mathbf{U}_\theta$ is then applied iteratively across $T$ timesteps, progressively reducing noise at each step to produce a cleaner intermediate image, $\mathbf{i}_{t-1}$, finally generating one of the \textit{cleanest} samples $\mathbf{i}_0$ from the target distribution.

The unconditional denoising diffusion could further be made \textit{conditional} by governing $\mathbf{U}_\theta$ with different auxiliary signals (\eg, textual captions) $\mathbf{c}$ \cite{rombach2022high}. Accordingly, $\mathbf{U}_\theta(\mathbf{i}_t,t, \mathbf{c})$ can denoise $\mathbf{i}_t$ while being influenced by $\mathbf{c}$, commonly via cross-attention \cite{rombach2022high}.

\label{sec:arch}
\noindent\textbf{Architecture.} \textit{Pixel-space} diffusion model \cite{ho2020denoising} is time-intensive as it operates on the full image resolution (\ie, $\mathbf{i}_0 \in \mathbb{R}^{h \times w \times 3}$). Contrarily, SD model \cite{rombach2022high} performs denoising in the \textit{latent-space}, resulting in significantly faster and more stable processing. In the {first} stage, SD trains a variational autoencoder (VAE) \cite{kingma2013auto} (an encoder $\mathbf{E}(\cdot)$ and a decoder $\mathbf{D}(\cdot)$). $\mathbf{E}(\cdot)$ transforms the input image into its latent representation $\mathbf{z}_0 =\mathbf{E}(\mathbf{i}_0) \in \mathbb{R}^{\frac{h}{8}\times \frac{w}{8}\times d}$. In the {second} stage, the UNet \cite{ronneberger2015u} $\mathbf{U}_\theta$, directly denoises the latent images. $\mathbf{U}_\theta$ comprises $12$ encoding, $1$ bottleneck, and $12$ \textit{skip-connected} decoding blocks \cite{rombach2022high}. Inside these encoding and decoding blocks, there are $4$ downsampling $(\mathbf{U}_{\mathbf{d}}^{1-4})$ and $4$ upsampling $(\mathbf{U}_{\mathbf{u}}^{1-4})$ layers respectively. In SD, a frozen CLIP textual encoder \cite{radford2021learning} $\mathbf{T}(\cdot)$ converts the textual prompt $\mathbf{c}$ into token-sequence, that influences the denoising process of $\mathbf{U}_\theta$ through cross-attention. $\mathbf{U}_\theta$ is trained over an $l_2$ objective as: $\mathcal{L}_{\text{l}_2}=\mathbb{E}_{\mathbf{z}_t,t,\mathbf{c},\epsilon} ({||\epsilon-\mathbf{U}_{\theta}(\mathbf{z}_t,t,\mathbf{T}(\mathbf{c}))||}_2^2)$. During inference, a noisy latent $\mathbf{z}_t$ is sampled directly as: $\mathbf{z}_t$$\sim$$\mathcal{N}(0,\mathbf{I})$. $\mathbf{U}_{\theta}$ removes noise from $\mathbf{z}_t$ iteratively over $T$ timesteps (conditioned on $\mathbf{c}$) to produce a denoised latent image $\hat{\mathbf{z}}_0\in \mathbb{R}^{\frac{h}{8}\times \frac{w}{8}\times d}$. The final high-resolution image is generated as: $\hat{\mathbf{i}}=\mathbf{D}(\hat{\mathbf{z}}_0)\in \mathbb{R}^{h\times w\times 3}$.

\vspace{-0.2cm}
\section{Pilot Study: Problems and Solution}
\vspace{-0.2cm}
\label{sec:pilot}
\noindent\textbf{What's Wrong with Diffusion Features?} \textit{Binary} and \textit{sparse} freehand sketches contain much lesser semantic contextual information than \textit{pixel-perfect} photos \cite{sangkloy2022sketch, chowdhury2023scenetrilogy, koley2023picture}. While SD \cite{rombach2022high} as a backbone feature extractor has established itself as a strong competitor in image-based vision tasks \cite{xu2023open, xu20233difftection, zhao2023unleashing, baranchuk2021label, de2023medical}, we dive deep into its internal representations to evaluate its effectiveness for the more challenging cross-modal sketch-based vision tasks. To this end, we ask the research question -- whether a frozen SD model is capable of extracting features from a sparse sketch that are \textit{equally-representative} as those derived from a \textit{pixel perfect} image. Given a few sketch-photo pairs, we extract the corresponding features (detail in \cref{sec:sd_feat}) from the second upsampling block of a frozen SD's UNet $\mathbf{U}_\theta$ with a fixed prompt $\mathtt{``a~photo~of~[CLASS]"}$ and $t=195$. We perform principal component analysis (PCA) on the extracted high-dimensional features and plot the first three principal components as RGB images for visualisation. As seen in \cref{fig:pilot} (left), features extracted from sketches are significantly inferior to those extracted from photos. Unlike photos, abstract freehand sketches are sparse and lack detailed textures/colours \cite{sangkloy2022sketch, chowdhury2023scenetrilogy, koley2023picture}, making it challenging for the SD model to capture \textit{fine-grained} visual information from it. We hypothesise that SD struggles to extract equally representative features from sketches due to their minimal semantic cues and high level of abstraction \cite{koley2023its}. Additionally, as the pretraining of SD was done predominantly on natural images \cite{schuhmann2022laion}, it is comparatively harder for SD to extract rich semantic information from a sparse sketch.

Digging deeper, we follow classical computer vision literature \cite{yin2019fourier, gonzalez2008digital} in shifting the focus from \textit{spatial}-domain, to analyse the behaviour of extracted features from a different perspective, \ie, exploring its nature in the \textit{frequency}-domain. We achieve this by calculating the 2D Fourier magnitude spectrum of the extracted features, where the central region represents the LF components, while the HF components are arranged radially outward from the centre \cite{gonzalez2008digital} (\cref{fig:pilot} {left}). Here we see that, although SD performs better on photos than sketches, it \textit{predominantly} focuses on HF components in \textit{both} cases.
This is likely because SD's UNet is primarily trained as a \textit{denoiser} thus operating with a \textit{global} emphasis \cite{zhang2024tale}, where its skip connections additionally propagate HF components (\eg, edges), suppressing its LF counterparts (\eg, overall semantic structures) to ease the reconstruction of high-res image from noise \cite{si2024freeu}. We hypothesise that this very fact is directly \textit{conflicting} with the core motivation of our dense prediction task from a \textit{sparse} data modality \ie, sketches, where capturing comprehensive semantic \cite{zhang2024tale} context (LF) takes much \textit{higher} precedence over preserving fine-details (HF), thus limiting SD’s effectiveness \cite{zhang2024tale, si2024freeu} for sketches, especially for dense prediction tasks (\eg, sketch-based segmentation).

\noindent\textbf{Reinforcing Diffusion Features.} We identify two key problems in SD feature extraction -- \textit{(i)} SD features are not equally-representative for sketches and photos, and \textit{(ii)} it mainly focuses on HF features, losing the crucial LF ones. In addressing these, our PCA analysis (\cref{fig:pilot} left) shows that, although SD is spatially-aware, it lacks accuracy. Inspired by existing works \cite{zhang2024tale} on hybrid models, we propose complementing the SD feature extractor by combining it with another pretrained model that offers high accuracy, even if it is less spatially-aware. To this end, we influence the SD feature extraction pipeline with a pretrained CLIP  \cite{radford2021learning} visual encoder as detailed in \cref{sec:methodology}. PCA maps of CLIP (\cref{fig:pilot} left) features depict highly accurate background-foreground separation, inherently providing the LF components that were missing in SD features. Finally, PCA maps of the proposed feature extractor (\cref{fig:pilot} left) demonstrate how the final representation is spatially-aware, accurate, integrates both HF and LF and works equally well for both sketches and photos. Apart from qualitative evidence, evaluating the performance of baseline SD+CLIP hybrid model, and the proposed feature extractor across different sketch-based discriminative and dense prediction tasks, we observe an average performance gain of $39.49\%$ (\cref{fig:pilot} {right}).

\vspace{-0.2cm}
\section{Proposed Methodology}
\vspace{-0.2cm}
\label{sec:methodology}
\noindent\textbf{Overview.} Observations from our pilot study (\cref{sec:pilot}) motivate us to reinforce the SD \cite{rombach2022high} feature extraction pipeline for better downstream task accuracy. However, reinforcing the SD feature extractor pipeline is non-trivial. Firstly, finetuning the entire model with task-specific dataset might interrupt SD's inherent pretrained knowledge \cite{koley2024text}. Furthermore, \textit{end-to-end} finetuning does not necessarily resolve the issues pertaining to spatial-awareness \vs accuracy or HF \vs LF trade-off (\cref{sec:pilot}). To this end, we resort to a hybrid model \cite{zhang2024tale} approach and use a complementary pretrained model (\ie, CLIP \cite{radford2021learning}) to strengthen the SD feature extractor pipeline. Secondly, obtaining features from CLIP alone is insufficient. They must be integrated effectively into the denoising process of the SD UNet to yield the performance gains observed in \cref{fig:pilot}. Finally, different layers of the SD's UNet capture a wide array of features across different scales and granularity (\ie, coarse to fine) \cite{tang2023emergent}, each with its own pros and cons suited to various downstream tasks \cite{koley2024text}. Thus, it is essential to effectively aggregate these multi-scale features to maximise their utility.

Specifically, our salient design choices are -- \textit{(i)} complementing SD \cite{rombach2022high} feature extraction with a pretrained CLIP \cite{radford2021learning} visual model, \textit{(ii)} an effective CLIP visual feature injection module, and \textit{(iii)} an aggregation module to unify features from different scales and granularities.

\vspace{-0.2cm}
\subsection{Model Architecture}
\vspace{-0.2cm}
\label{sec:sd_feat}
\noindent\textbf{Stable Diffusion Feature Extraction.} Internal activation maps of a pretrained SD UNet $\mathbf{U}_\theta$ contain information-rich representations that have found use in tasks like image segmentation \cite{baranchuk2021label}, object detection \cite{xu20233difftection}, classification \cite{li2023your}, etc. Here, we aim to use these internal representations as an efficient backbone feature for various sketch-based vision tasks. Given an image-text pair $(\mathbf{i}, \mathbf{c})$, the input image is first passed through the pretrained VAE \cite{kingma2013auto} encoder $\mathbf{E}(\cdot)$, converting it into an initial latent representation $\mathbf{z}_0=\mathbf{E}(\mathbf{i}) \in \mathbb{R}^{\frac{h}{8}\times \frac{w}{8}\times d}$. Then, given a timestep value $t$, SD performs forward diffusion on $\mathbf{z}_0$ to convert it to the $t^{th}$ step noisy latent $\mathbf{z}_t$. Finally, the textual prompt embedding (\cref{sec:arch}) $\mathbf{T}(\mathbf{c})$, noisy latent $\mathbf{z}_t$, and timestep $t$ are passed to the pretrained denoising UNet \cite{rombach2022high} $\mathbf{U}_\theta(\cdot)$ to extract the internal features from its \textit{skip-connected} upsampling layers $\{\mathbf{U}_{\mathbf{u}}^{n} \rightarrow f_{\mathbf{u}}^{n}\}_{n=1}^{4}$. In SD v2.1 \cite{rombach2022high}, for an input image $\mathbf{i} \in \mathbb{R}^{h \times w \times 3}$, upsampling layers $\{\mathbf{U}_{\mathbf{u}}^{n}\}_{n=1}^{4}$ generate features as: $f_{\mathbf{u}}^{1} \in \mathbb{R}^{\frac{h}{32} \times \frac{w}{32} \times 1280}$, $f_{\mathbf{u}}^{2} \in \mathbb{R}^{\frac{h}{16} \times \frac{w}{16} \times 1280}$, $f_{\mathbf{u}}^{3} \in \mathbb{R}^{\frac{h}{8} \times \frac{w}{8} \times 640}$, and $f_{\mathbf{u}}^{4} \in \mathbb{R}^{\frac{h}{8} \times \frac{w}{8} \times 320}$ respectively. Due to the unavailability of paired textual prompts in most sketch-photo datasets \cite{sangkloy2016the, ha2017neural, eitz2012humans}, we use \textit{null} (\ie, $\mathtt{``~"}$) prompt in all our experiments.

\vspace{-0.2cm}
\begin{figure}[!htbp]
    \centering
    \includegraphics[width=1\linewidth]{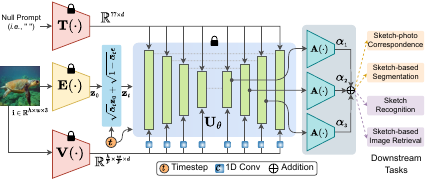}
    \vspace{-0.6cm}
    \caption{Given the frozen SD \cite{rombach2022high} and CLIP \cite{radford2021learning} models, the proposed method learns the \textit{aggregation network}, $1D$ \textit{convolutional layers}, and \textit{branch-weights} with sketch-photo pairs, via different losses for different downstream tasks (details in \cref{sec:exp}).}
    \label{fig:arch}
    \vspace{-0.2cm}
\end{figure}

\label{sec:clip}
\noindent\textbf{CLIP Visual Feature Extraction.} Akin to SD, pretrained CLIP \cite{radford2021learning} encoders have also been used in a plethora of vision tasks like segmentation \cite{mukhoti2023open}, retrieval \cite{sain2023clip}, object detection \cite{chowdhury2023what}, video captioning \cite{yang2023vid2seq}, to name a few. In this work, we aim to influence the intermediate feature maps of SD \cite{rombach2022high} with that from the CLIP visual encoder. Specifically, the input image $\mathbf{i}$ is passed through the pretrained CLIP visual encoder $\mathbf{V}(\cdot)$ to extract the patched feature $f_{\mathbf{v}}$ from the penultimate layer as: $f_{\mathbf{v}} = \mathbf{V}(\mathbf{i}) \in \mathbb{R}^{\frac{h}{p} \times \frac{w}{p} \times d}$, where $p$ is the patch-size of the $\mathbf{V}(\cdot)$ backbone encoder.

\noindent\textbf{CLIP Feature Injection.} Based on the qualitative and quantitative evidence presented in the pilot study (\cref{sec:pilot}), we hypothesise that CLIP \cite{radford2021learning} visual and textual embeddings are intrinsically aligned \cite{radford2021learning, sangkloy2022sketch}. Prior works \cite{sangkloy2022sketch} have underscored the importance of textual prompts in the quality of extracted features \cite{sangkloy2022sketch}. However, most simple prompts (\eg, $\mathtt{``a~photo/sketch~of~[CLASS]"}$) are inherently \textit{coarse-grained}, offering {only} a \textit{high-level} description of an image. Such general prompts often lack the semantic detail necessary to fully characterise an image \cite{chowdhury2023scenetrilogy, koley2023its}, limiting the model’s ability to generate highly accurate and contextually-rich feature representations \cite{koley2024text}.

To address this, we propose injecting CLIP’s visual features into multiple layers of the SD's UNet, thereby influencing the denoising process at various stages. This approach not only enhances the accuracy of SD features but also enables the model to draw on semantic information from both CLIP's visual and textual embeddings.

Specifically, after extracting CLIP's visual embeddings $f_\mathbf{v} \in \mathbb{R}^{\frac{h}{p} \times \frac{w}{p} \times d}$, we pass them through simple learnable $1D$ convolutional layers $\mathcal{C}(\cdot)$ to modify them \textit{just enough} to match the feature dimension of different blocks of the UNet. These transformed feature maps are then added to the intermediate UNet features $\{f_{\mathbf{u}}^{n}\}_{n=1}^{4}$ at every timestep of the denoising process as: $\hat{f}_{\mathbf{u}}^{n} = f_{\mathbf{u}}^{n} + \mathcal{C}(f_\mathbf{v})$. This injection occurs in all timesteps and layers of the UNet, enabling the model to influence the denoising process with the semantic information encoded within CLIP's visual embeddings.

\noindent\textbf{SD Feature Aggregation.} Different layers of $\mathbf{U}_\theta$ capture features of varying \textit{semantic-granularity} \cite{tang2023emergent}, where features from specific layers are ideally suited for certain downstream tasks \cite{koley2024text}.
Instead of manually selecting each layer for a specific task, we introduce a \textit{dynamic} approach, allowing the model to automatically extract the most optimal features, thus eliminating the need for manual tuning.

To achieve this, we train a feature aggregation network comprising three ResNet \cite{he2016deep} blocks $\mathbf{A}(\cdot)$, which transform the CLIP-enhanced SD features $\{\hat{f}_{\mathbf{u}}^{n}\}_{n=1}^{3}$ extracted from the first three UNet upsampling layers as: $\bar{f}_{\mathbf{u}}^{n}=\mathbf{A}(\hat{f}_{\mathbf{u}}^{n}) \in \mathbb{R}^{60\times 60\times d}$. The final feature map is then obtained by computing a \textit{weighted summation} of the aggregated features using a learnable weight parameter $\{\alpha_{n}\}_{n=1}^{3}$ for each feature-branch. This aggregation strategy enables the model to \textit{dynamically} adjust its focus on different layers, adaptively modulating the contribution of each feature-branch based on the \textit{semantic-granularity} of the task.

\vspace{-0.2cm}
\section{Experiments}
\vspace{-0.2cm}
\label{sec:exp}
The proposed feature extractor enriched with the large-scale pretrained knowledge of SD \cite{rombach2022high} and CLIP \cite{radford2021learning}, enables multiple sketch-based \textit{discriminative} and \textit{dense prediction} tasks. Specifically, we showcase its efficacy in tasks like -- sketch-based image retrieval \cite{sain2023clip}, sketch recognition \cite{yang2021sketchaa}, sketch-photo correspondence learning \cite{lu2023learning}, and sketch-based image segmentation \cite{hu2020sketch} on different datasets \cite{sangkloy2016the, eitz2012humans, lu2023learning, ha2017neural} and scenarios. In all cases, we \textit{only} learn $\mathcal{C}(\cdot)$, aggregation network $\mathbf{A}(\cdot)$, and branch-weights $\alpha_n$, keeping pretrained SD, CLIP models frozen.

\vspace{-0.2cm}
\subsection{Sketch-based Image Retrieval (SBIR)}
\vspace{-0.2cm}
\noindent\textbf{Problem Statement.} Given a query sketch $\mathbf{s}\in\mathbb{R}^{h\times w\times 3}$ of any class `\textit{j}', \textit{category-level} SBIR aims to retrieve a photo $\mathbf{p}_i^j$ from the \textit{same} class, out of a gallery $\mathcal{G}=\{\mathbf{p}_i^j\}_{i=1}^{N_j}{|}_{j=1}^{N_c}$ containing images from a total $N_c$ classes with $N_j$ images per class \cite{yelamarthi2018zero}. Whereas, \textit{cross-category fine-grained} (FG) SBIR setup intends to learn a unified model capable of \textit{instance-level} matching from a gallery containing photos from \textit{multiple} ($N_c$) classes. Conventional SBIR is evaluated on categories $\mathcal{C}^S=\{c_1^S, \cdots, c_{N}^S\}$ that are \textit{seen} during training. Contrarily, the zero-shot (ZS) paradigm \cite{sain2023clip} focuses on evaluating on \textit{mutually exclusive} (\ie, $\mathcal{C}^S \cap \mathcal{C}^U=\emptyset$) \textit{unseen} categories $\mathcal{C}^U=\{c_1^U, \cdots, c_{M}^U\}$.

\noindent\textbf{Training \& Evaluation.} Following standard literature \cite{sain2023clip}, we use triplet loss to train our SBIR models. Specifically, given the global max-pooled features from the proposed feature extractor, triplet loss aims to minimise the distance $\delta$ between an anchor sketch ($\mathbf{s}$) feature $f_\mathbf{s} \in \mathbb{R}^{d}$ and a positive photo ($\mathbf{p}$) feature $f_\mathbf{p}\in \mathbb{R}^{d}$ from the \textit{same} category as $\mathbf{s}$, while maximising the same from a negative photo ($\mathbf{n}$) feature $f_\mathbf{n}\in \mathbb{R}^{d}$ of a \textit{different} category. Whereas, in case of \textit{cross-category} FG-SBIR framework we use \textit{hard} triplets, where the negative sample is a \textit{distinct instance} ($\mathbf{p}^j_k;k\neq i$) belonging to the \textit{same class} as the anchor sketch ($\mathbf{s}_i^j$) and its paired photo ($\mathbf{p}_i^j$) \cite{koley2024text}. With margin $\mu$, triplet loss becomes:

\vspace{-0.4cm}
\begin{small}
\begin{equation}
    \mathcal{L}_\mathrm{triplet}(f_\mathbf{s},f_\mathbf{p},f_\mathbf{n}) = \mathtt{max}\{0,\mu+\delta(f_\mathbf{s},f_\mathbf{p})-\delta(f_\mathbf{s},f_\mathbf{n})\} \quad .
\end{equation}
\end{small}

\noindent For category-level ZS-SBIR, following~\cite{sain2023clip, koley2024text}, we consider top $k$ retrieved images to calculate mean average precision (mAP@k) and precision (P@k). Whereas, for cross-category ZS-FG-SBIR, we measure Acc.@k, which indicates the percentage of sketches having true-matched photos among the top-k retrieved images.

\noindent\textbf{Dataset \& Competitors.} Here we use three datasets -- \textit{(i)} \textbf{Sketchy} \cite{sangkloy2016the} and \textbf{Sketchy-extended} \cite{liu2017deep, sangkloy2016the} datasets are used to train and test our cross-category ZS-FG-SBIR and category-level ZS-SBIR models respectively. While Sketchy \cite{sangkloy2016the} holds $12,500$ photos from $125$ classes, each having at least $5$ fine-grained paired sketches, its extended version \cite{liu2017deep} carries additional $60,652$ photos from ImageNet \cite{deng2009imagenet}. We use photos/sketches from $104$ classes for training and $21$ for testing \cite{yelamarthi2018zero}. \textit{(ii)} \textbf{Tu-Berlin} \cite{eitz2012humans} contains $204,489$ photos from $250$ classes, each having $\sim 80$ sketches. We use photos/sketches from $220$ classes for training and $30$ for testing. \textit{(iii)} \textbf{Quick, Draw!} \cite{ha2017neural} holds over $50M$ sketches from $345$ categories. We use photos/sketches from $80$ classes for training and $30$ for testing.

For category-level ZS-SBIR, we compare our method with SoTA methods like \textit{SAKE} \cite{liu2019semantic}, \textit{CAAE} \cite{yelamarthi2018zero}, \textit{CCGAN} \cite{dutta2019semantically}, \textit{GRL} \cite{dey2019doodle}, \textit{SD-PL} \cite{koley2024text}, etc. Whereas, for cross-category ZS-FG-SBIR, we compare with \textit{Gen-VAE} \cite{pang2019generalising}, \textit{LVM} \cite{sain2023clip}, and \textit{SD-PL} \cite{koley2024text}. Apart from the SoTA methods, we also compare \textit{all} tasks against a few self-designed \textit{B}aselines. Among them \textit{B-CLIP}, \textit{B-DINO}, \textit{B-DINOv2}, and \textit{B-SD} use \textit{off-the-shelf} pretrained CLIP \cite{radford2021learning} ViT-L/14, DINO \cite{caron2021emerging} ViT-S/8, DINOv2 \cite{oquab2023dinov2} ViT-S/14, or SD \cite{rombach2022high} v2.1 encoders as feature extractors, without task-specific training. {We also form another baseline \textit{B-SD+CLIP} as a straightforward combination of SD and CLIP features without the proposed feature injection and aggregation modules. Finally, \textit{B-Finetuning} finetunes the \textit{entire} pretrained SD UNet and CLIP visual encoder on task-specific data.}

\begin{table}[t]
    \centering
    \resizebox{\columnwidth}{!}{
    \begin{tabular}{lcccccc}
    \toprule
        \multicolumn{1}{c}{\multirow{2}{*}{\textbf{Methods}}} & \multicolumn{2}{c}{\textbf{Sketchy}~\cite{sangkloy2016the}} & \multicolumn{2}{c}{\textbf{TU-Berlin}~\cite{eitz2012humans}} & \multicolumn{2}{c}{\textbf{Quick, Draw!}~\cite{ha2017neural}} \\\cmidrule(lr){2-3}\cmidrule(lr){4-5}\cmidrule(lr){6-7}
        & \textbf{mAP@200} & \textbf{P@200} & \textbf{mAP@all} & \textbf{P@100} & \textbf{mAP@all} & \textbf{P@200}  \\\cmidrule(lr){1-3}\cmidrule(lr){4-5}\cmidrule(lr){6-7}
        B-CLIP                           & 0.250 & 0.261 & 0.228 & 0.257 & 0.080 & 0.141\\
        B-DINO                           & 0.493 & 0.481 & 0.450 & 0.492 & 0.167 & 0.249\\
        B-DINOv2                         & 0.527 & 0.533 & 0.481 & 0.532 & 0.170 & 0.268\\
        B-SD                             & 0.558 & 0.571 & 0.510 & 0.561 & 0.179 & 0.287\\
        B-SD+CLIP                        & 0.588 & 0.592 & 0.537 & 0.589 & 0.179 & 0.311\\
        B-Finetuning                     & 0.120 & 0.172 & 0.011 & 0.010 & 0.003 & 0.006 \\
        \cmidrule(lr){1-3}\cmidrule(lr){4-5}\cmidrule(lr){6-7}
        SAKE~\cite{liu2019semantic}             & 0.497 & 0.598 & 0.475 & 0.599 & --    & --\\
        IIAE~\cite{hwang2020variational}        & 0.373 & 0.485 & 0.412 & 0.503 & --    & --\\
        CAAE~\cite{yelamarthi2018zero}          & 0.156 & 0.260 & 0.005 & 0.003 & --    & --\\
        CCGAN~\cite{dutta2019semantically}      & --    & --    & 0.297 & 0.426 & --    & --\\
        CVAE~\cite{yelamarthi2018zero}          & 0.225 & 0.333 & 0.005 & 0.001 & 0.003 & 0.003\\
        GRL~\cite{dey2019doodle}                & 0.369 & 0.370 & 0.110 & 0.121 & 0.075 & 0.068\\
        LVM~\cite{sain2023clip}                 & 0.723 & 0.725 & 0.651 & 0.732 & 0.202 & 0.388\\
        SD-PL \cite{koley2024text}              & 0.746 & 0.747 & 0.680 & 0.744 & 0.231 & 0.397\\
        \rowcolor{MidnightBlue!30}
        \textbf{\textit{Proposed}}              & \bf0.761 & \bf0.763 & \bf0.695 & \bf0.753 & \bf0.242 & \bf0.399\\
        \bottomrule
    \end{tabular}
    }
      \vspace{-0.3cm}
        \caption{Results for \textit{category-level ZS-SBIR}.}
    \label{tab:zs-sbir}
    \vspace{-0.7cm}
\end{table}

\vspace{-0.2cm}
\begin{table}[!htbp]
    \centering
    \resizebox{\columnwidth}{!}{
    \begin{tabular}{lcc|lcc}
    \toprule
     \multicolumn{1}{c}{\multirow{1}{*}{\textbf{Methods}}}  & \textbf{Acc.@1} & \textbf{Acc.@5} & \multicolumn{1}{c}{\multirow{1}{*}{\textbf{Methods}}} & \textbf{Acc.@1} & \textbf{Acc.@5}\\\cmidrule(lr){1-3}\cmidrule(lr){4-6}

         B-CLIP       & 11.84 & 21.66 &  B-Finetuning                           & 5.67 & 9.17\\
         B-DINO       & 22.49 & 46.97 &  Gen-VAE \cite{pang2019generalising}    & 22.60 & 49.00\\
         B-DINOv2     & 21.19 & 44.31 &  LVM \cite{sain2023clip}                & 28.68 & 62.34\\
         B-SD         & 23.98 & 49.42 &  SD-PL \cite{koley2024text}             & 31.94 & 65.81\\
         B-SD+CLIP    & 24.16 & 52.01 & \cellcolor{MidnightBlue!30}\textbf{\textit{Proposed}}  & \cellcolor{MidnightBlue!30} \bf33.01 & \cellcolor{MidnightBlue!30} \bf67.92\\
            \bottomrule

    \end{tabular}
    }
        \vspace{-0.3cm}
        \caption{Results on Sketchy \cite{sangkloy2016the} for \textit{cross-category ZS-FG-SBIR}.}

        \label{tab:cc-fg-zs-sbir}
    \vspace{-0.2cm}
\end{table}

\vspace{-0.4cm}
\subsection{Sketch Recognition}
\vspace{-0.2cm}
\noindent\textbf{Problem Statement.} Sketch recognition deals with identifying and classifying freehand abstract sketches into predefined categories \cite{xu2018sketchmate, ribeiro2020sketchformer}. Given a dataset $\mathcal{D}=\{(\mathbf{s}_i,\mathbf{y}_i)\}_{i=1}^{N}$  containing $N$ sketches and their corresponding labels from $C$ classes, the goal is to learn a model that can correctly recognise the class of an input query sketch.

\noindent\textbf{Training \& Evaluation.} Given the global max-pooled query sketch feature $f_\mathbf{s} \in \mathbb{R}^{d}$, we use standard cross-entropy (CE) loss to train our sketch recognition model as,

\vspace{-0.3cm}
\begin{small}
\begin{equation}
    \mathcal{L}_\mathrm{CE}(\mathbf{s}_i,\mathbf{y}_i) = -\frac{1}{N}\sum_{i=1}^{N}\text{log}~p(\mathbf{y}_i|\mathbf{s}_i) \quad .
\end{equation}
\end{small}
\vspace{-0.3cm}

\noindent Following \cite{xu2018sketchmate, yang2021sketchgnn, qu2023sketchxai} we evaluate using Acc.@k.

\noindent\textbf{Dataset \& Competitors.} For this task, we again use the popular \textbf{TU-Berlin} \cite{eitz2012humans} and \textbf{Quick, Draw!} \cite{ha2017neural} datasets. For Quick, Draw!, we adopt the data split from \cite{ha2017neural}, with each class containing $70,000$, $2,500$, and $2500$ training, validation, and test samples. Apart from the self-designed baselines, we also compare against SoTA methods like \textit{SketchMate} \cite{xu2018sketchmate}, \textit{SketchGNN} \cite{yang2021sketchgnn}, and \textit{SketchXAI} \cite{qu2023sketchxai}.

\begin{table}[t]
    \centering
    \resizebox{\columnwidth}{!}{
    \begin{tabular}{lcc|lcc}
    \toprule
     \multicolumn{1}{c}{\multirow{1}{*}{\textbf{Methods}}}  & \textbf{TU-Berlin} & \textbf{Quick, Draw!} & \multicolumn{1}{c}{\multirow{1}{*}{\textbf{Methods}}} & \textbf{TU-Berlin} & \textbf{Quick, Draw!} \\\cmidrule(lr){1-3}\cmidrule(lr){4-6}

         B-CLIP       & 30.09 & 30.87 &  B-Finetuning                          & 10.29 & 12.37\\
         B-DINO       & 51.79 & 53.28 &  SketchMate \cite{xu2018sketchmate}    & 77.96 & 79.44 \\
         B-DINOv2     & 58.01 & 59.12 &  SketchGNN \cite{yang2021sketchgnn}    & 76.43 & 77.31 \\
         B-SD         & 61.37 & 63.57 &  SketchXAI \cite{qu2023sketchxai}      &  --   & 86.10 \\
         B-SD+CLIP    & 65.47 & 67.78 & \cellcolor{MidnightBlue!30}\textbf{\textit{Proposed}}  & \cellcolor{MidnightBlue!30} \bf84.96 & \cellcolor{MidnightBlue!30} \bf87.02 \\
            \bottomrule

    \end{tabular}
    }
        \vspace{-0.3cm}
        \caption{Acc.@1 results for \textit{sketch recognition}.}

        \label{tab:recognition}
    \vspace{-0.7cm}
\end{table}

\vspace{-0.2cm}
\subsection{Sketch-photo Correspondence}
\vspace{-0.2cm}
\noindent\textbf{Problem Statement.} Sketch-photo correspondence learning involves identifying a mapping between a freehand sketch of an object and its corresponding photo, such that keypoints in the sketch can be semantically aligned with corresponding keypoints in the photo \cite{lu2023learning}. Given a sketch-photo pair $\{\mathbf{s},\mathbf{p}\}\in\mathbb{R}^{h\times w\times 3}$ and a set of $N$ query keypoints (on sketch) $K_\mathbf{s}=\{x_i^s, y_i^s\}_{i=1}^N$, the aim is to predict a new set of corresponding keypoints $\hat{K}_\mathbf{p}=\{\hat{x}_i^\mathbf{p}, \hat{y}_i^\mathbf{p}\}_{i=1}^N$ on the photo, such that $\hat{K}_\mathbf{p}$ aligns \textit{semantically} with $K_\mathbf{s}$.

\noindent\textbf{Training \& Evaluation.} Following \cite{zhang2024telling}, we use CLIP-style contrastive \cite{radford2021learning} and end-point error losses to train our sketch-photo correspondence model. Given the sketch and photo features $\{f_\mathbf{s}, f_\mathbf{p}\} \in\mathbb{R}^{60\times 60\times d}$ from the proposed feature extractor, contrastive loss ($\mathcal{L}_{\text{CL}}$) aims to align the \textit{individual} patches ($\mathbb{R}^{d}$) containing each of the paired keypoints from both sketch and photo. Whereas for end-point error loss, we first estimate optical flows \cite{zhang2024telling} from the extracted features, followed by calculating $l_2$ norm between the estimated ($\mathbf{f}_e$) and ground truth ($\mathbf{f}_g$) flow \cite{zhang2024telling} as:

\vspace{-0.2cm}
\begin{small}
\begin{equation}
    \mathcal{L}_{\text{EPE}}(\mathbf{f}_e^i,\mathbf{f}_g^i) = \frac{1}{M} \sum_{i=1}^{M}||\mathbf{f}_e^i-\mathbf{f}_g^i||_2^2  \quad ,
\end{equation}
\end{small}
\vspace{-0.2cm}

\noindent where, $M$ denotes the patch number (\ie, $60$). Finally, the overall loss becomes $\mathcal{L}_{\text{corr}}=\mathcal{L}_{\text{CL}}$ + $\mathcal{L}_{\text{EPE}}$. Following \cite{lu2023learning} we use PCK@k as a test metric which measures the percentage of \textit{k}eypoints \textit{p}redicted \textit{c}orrectly within a distance threshold (\ie, $k\%$ of the image size).

\noindent\textbf{Dataset \& Competitors.} For this task, we use the \textbf{PSC6K} \cite{lu2023learning} dataset that contains more than $150K$ keypoint annotations for $1250$ photos, each having $5$ paired sketches (taken from the original Sketchy \cite{sangkloy2016the} test split) from $125$ classes. We use a $60$:$40$ split for training and testing, where we compare the proposed approach against the self-designed baselines and SoTA methods like \textit{WeakAlign}\cite{rocco2018end}, \textit{WarpC} \cite{truong2021warp}, \textit{Self-Supervised} \cite{lu2023learning}, etc.

\vspace{-2mm}
\begin{table}[!htbp]
\centering
\footnotesize
\resizebox{\columnwidth}{!}{
\begin{tabular}{lcc|lcc}
\toprule
\multicolumn{1}{c}{\textbf{Methods}} & \textbf{PCK@5} & \textbf{PCK@10} & \multicolumn{1}{c}{\textbf{Methods}} &\textbf{PCK@5} & \textbf{PCK@10}  \\ \cmidrule(lr){1-3}\cmidrule(lr){4-6}
B-CLIP              & 22.29 & 30.41 & B-Finetuning                    & 11.58 & 17.69\\
B-DINO              & 34.21 & 48.59 & WeakAlign \cite{rocco2018end}   & 43.55 & 78.60\\
B-DINOv2            & 39.91 & 56.46 & WarpC \cite{truong2021warp}     & 56.78 & 79.70\\
B-SD                & 41.42 & 58.71 & Self-Sup. \cite{lu2023learning} & 58.00 & 84.93 \\
B-SD+CLIP           & 44.90 & 62.02 &\cellcolor{MidnightBlue!30}\textbf{\textit{Proposed}} & \cellcolor{MidnightBlue!30}\textbf{70.31} & \cellcolor{MidnightBlue!30}\textbf{89.86} \\
\bottomrule
\end{tabular}
}
\vspace{-0.3cm}
\caption{Results on PSC6K \cite{lu2023learning} for \textit{sketch-photo correspondence}.}

\label{table:corr}
\vspace{-0.5cm}
\end{table}

\vspace{-0.3cm}
\subsection{Sketch-based Image Segmentation}
\vspace{-0.2cm}
\noindent\textbf{Problem Statement.} Given a query sketch $\mathbf{s}\in\mathbb{R}^{h\times w\times 3}$ of any catgory, sketch-based image segmentation \cite{hu2020sketch} aims to predict a binary segmentation mask, $\mathbf{m}\in \mathbb{R}^{h\times w \times 1}$, depicting the pixel-locations, wherever the \textit{queried-concept} appears in \textit{any} candidate image $\mathbf{p}\in\mathbb{R}^{h\times w\times 3}$ of \textit{that} category, from an \textit{entire} test gallery. The predicted mask marks pixels belonging to that \textit{sketched-concept} as $1$ and the rest as $0$.

\noindent\textbf{Training \& Evaluation.} We train our segmentation model using binary cross-entropy (BCE) loss, between predicted and ground truth (GT) masks. Given \textit{patched} photo feature $f_\mathbf{p} \in\mathbb{R}^{60\times 60\times d}$ and \textit{global max-pooled} sketch feature photo features $f_\mathbf{s} \in\mathbb{R}^{d}$ from the proposed feature extractor, we calculate the cosine similarity between them to generate a \textit{patch-wise} correlation map $\mathbf{c} = f_\mathbf{s}\odot f_\mathbf{p}\in\mathbb{R}^{60\times60}$. We upsample $\mathbf{c}$ to the original spatial resolution (\ie, $\mathbb{R}^{h\times w}$), followed by \textit{differentiable} Sigmoid thresholding \cite{chen2021localizing} to get the predicted map. BCE loss between the GT and predicted maps can be calculated as,
\begin{equation}
\footnotesize
\vspace{-0.1cm}
\mathcal{L}_{\text{BCE}}(y_i, \hat{y}_i) = -\frac{1}{N} \sum_{i=1}^{N} \left[ y_i \log(\hat{y}_i) + (1 - y_i) \log(1 - \hat{y}_i) \right] \quad ,
\vspace{-0.1cm}
\end{equation}

\noindent where, $N$ is the number of total pixels, $y_i$ and $\hat{y}_i$ are the GT and predicted mask pixels. During testing, we calculate similar \textit{patch-wise} correlation map from sketch/photo features and upsample it to the original spatial image dimension, followed by thresholding it with an empirically chosen fixed value ($0.47$) and standard CRF \cite{krahenbuhl2011efficient}-based post-processing. We use mean Intersection over Union (mIoU) and pixel accuracy (pAcc.) as test metrics, which measure the overlap and percentage of correctly predicted pixels between predicted and GT segmentation masks respectively.

\noindent\textbf{Dataset \& Competitors.} Due to the lack of existing paired sketch-photo-mask datasets, we manually annotate photos of $10$ chosen classes from the Sketchy \cite{sangkloy2016the} dataset with binary mask annotations, resulting in $5K$ paired sketch-photo-mask tuples. We split it into $80$:$20$ for training and testing. Besides baselines and SoTA \textit{Sketch-a-Segmenter} \cite{hu2020sketch}, we also compare with a modified version of \textit{ZS-Seg} \cite{xu2022simple}, which adapts a simple text-based zero-shot segmentation framework for sketches by replacing its CLIP \cite{radford2021learning} textual encoder with the visual one to encode input sketches.

\vspace{-0.2cm}
\begin{figure}[h]
    \centering
    \includegraphics[width=1\linewidth]{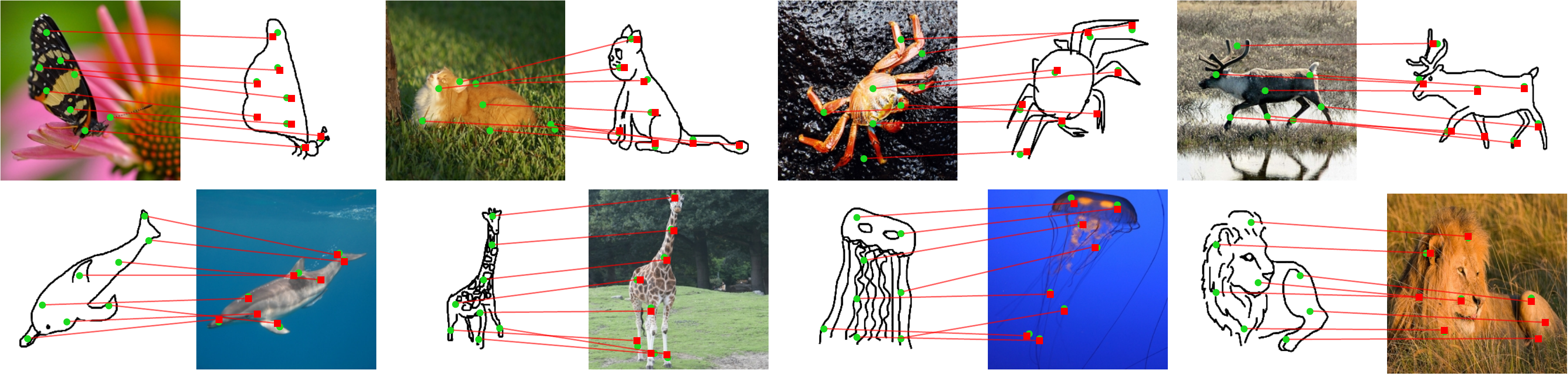}
    \vspace{-0.8cm}
    \caption{Sketch-photo correspondence (left$\shortrightarrow$right : source $\shortrightarrow$ target) results on PSC6K. \green{Green} circles and squares depict source and GT points respectively, while \red{red} squares denote predicted points.}
    \label{fig:corr}
    \vspace{-0.2cm}
\end{figure}

\vspace{-0.3cm}
\subsection{Results and Analyses}
\begin{figure}[!htbp]
    \centering
    \includegraphics[width=1\linewidth]{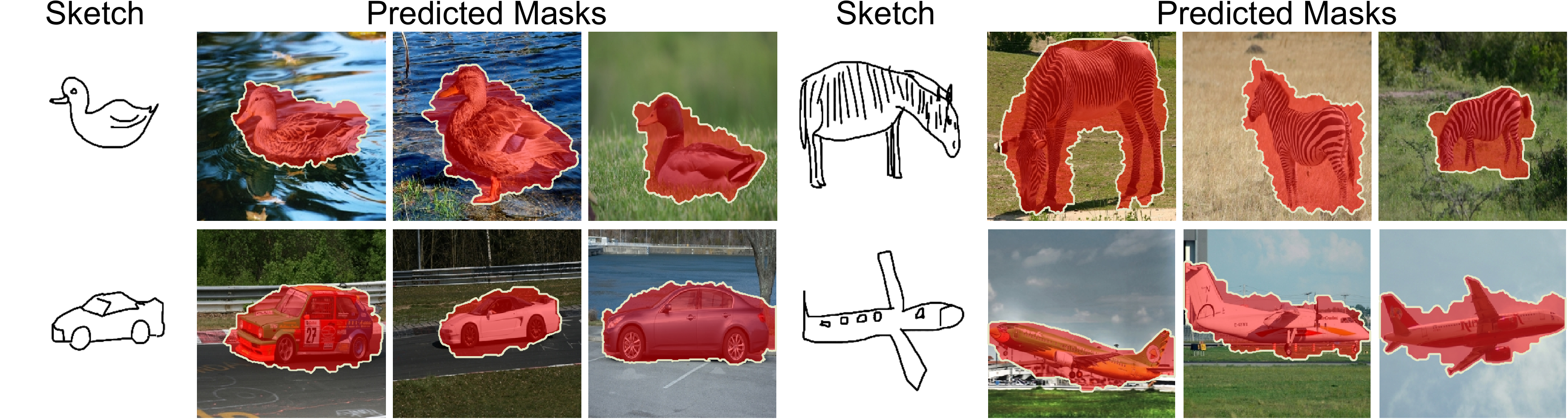}
    \vspace{-0.6cm}
    \caption{Qualitative results for sketch-based image segmentation. Given a query sketch, our method generates separate segmentation masks for \textit{all} images of that category. \textit{(Zoom-in for the best view.)}}
    \label{fig:seg}
    \vspace{-0.6cm}
\end{figure}

\vspace{-0.2cm}
Following are the observations from quantitative and qualitative results presented in Tab.\ \ref{tab:zs-sbir}-\ref{tab:segmentation} and Fig.\ \ref{fig:corr}-\ref{fig:seg}. \textit{(i)} Naive adaptation of pretrained backbones (\eg, B-CLIP, B-DINO, B-SD, etc.) fail to beat our method in both ZS-SBIR and ZS-FG-SBIR settings (Tab.\ \ref{tab:zs-sbir}-\ref{tab:cc-fg-zs-sbir}), where we achieve an average $76.15~(54.32)\%$ higher mAP@200 (Acc.@1) over all competitors on Sketchy. We posit that our strategic combination of SD and CLIP brings the \textit{``best of both worlds''}, complementing each other's flaws. \textit{(ii)} While SoTA sketch-recognition frameworks \cite{xu2018sketchmate, yang2021sketchgnn} perform reasonably across different datasets (\cref{tab:recognition}), our method exceeds them with an average Acc.@1 gain of $48.09\%$ (Quick, Draw!). SketchXAI \cite{qu2023sketchxai} despite being our strongest contender, depicts sub-optimal Acc.@1. This gain is likely due to the large-scale pretraining of SD and CLIP, which enables superior \textit{open-world} object understanding \cite{zhang2024tale}. \textit{(iii)} For sketch-photo correspondence learning, our method shows an impressive $21.22\%$ PCK@5 gain over SoTA \cite{lu2023learning}, without the time-consuming two-stage training of \cite{lu2023learning}, or complicated warp-estimation of \cite{truong2021warp}, thanks to {SD's innate \textit{object-localisation} capability}. \textit{(iv)} Sketch-based segmentation being the most challenging task amongst all, baseline methods perform quite poorly (\cref{tab:segmentation}). However, the proposed method surpasses SoTA \cite{hu2020sketch} with a mIoU (pAcc.) margin of $29.42~(24.27)\%$\cut{, likely due to our stronger backbone}. \textit{(v)} B-Finetuning fails drastically in almost all tasks (Tab.\ \ref{tab:zs-sbir}-\ref{tab:segmentation}). We hypothesise that finetuning entire SD and CLIP models with \textit{limited} sketch data distorts their large-scale pretrained knowledge \cite{sain2023clip, koley2024text}, thus sacrificing the \textit{generalisation potential}. \textit{(vi)} Finally, B-SD+CLIP depicts a $\sim$$35.78\%$ lesser scores than ours (averaged over all tasks), which again underpins the need for the proposed feature injection and aggregation strategy, rather than a simple combination of pretrained foundation models.

\begin{table}[t]
    \centering
    \resizebox{\columnwidth}{!}{
    \begin{tabular}{lcc|lcc}
    \toprule
     \multicolumn{1}{c}{\multirow{1}{*}{\textbf{Methods}}}  & \textbf{mIoU} & \textbf{pAcc.} & \multicolumn{1}{c}{\multirow{1}{*}{\textbf{Methods}}} & \textbf{mIoU} & \textbf{pAcc.} \\\cmidrule(lr){1-3}\cmidrule(lr){4-6}

 B-CLIP       & 20.63 & 26.82 &  B-Finetuning                            & 20.36 & 21.84\\
 B-DINO       & 32.81 & 42.06 &  DeepLabv3+Sketch \cite{hu2020sketch}    & 40.63 & 55.23 \\
 B-DINOv2     & 34.17 & 45.80 &  ZS-Seg \cite{xu2022simple}              & 44.73 & 60.97\\
 B-SD         & 39.02 & 49.03 &  Sketch-a-Segmenter \cite{hu2020sketch}  & 46.45 & 60.28\\
 B-SD+CLIP    & 41.87 & 51.91 &  \cellcolor{MidnightBlue!30}\textbf{\textit{Proposed}}  & \cellcolor{MidnightBlue!30} \bf60.12& \cellcolor{MidnightBlue!30} \bf74.91\\\bottomrule

    \end{tabular}
    }
        \vspace{-0.3cm}
    \caption{Results for \textit{sketch-based image segmentation}.}

        \label{tab:segmentation}
    \vspace{-0.7cm}
\end{table}

\vspace{-0.2cm}

\vspace{-0.2cm}
\section{Ablation on Design}
\vspace{-0.2cm}
\label{sec:abal}

\noindent\textbf{{[i]} Effect of aggregation network.}
To judge its efficacy, instead of aggregation networks $\mathbf{A}(\cdot)$, we manually interpolate features from $\{\mathbf{U}_{\mathbf{u}}^n\}_{n=1}^{3}$ into $\mathbb{R}^{60\times 60\times d}$ dimension and add them (via $\alpha_n$) to form the final feature map. A rapid PCK@5 (mIoU) drop of $28.65~(26.51)\%$, in case of \textbf{w/o Aggregation Network} (\cref{tab:abal}) indicates its significance in extracting semantically-meaningful features.

\noindent\textbf{{[ii]} Are learnable branch weights necessary?}
Learnable weights dynamically ensure the \textit{optimum importance} of each feature-branch for a specific task. Although less pronounced in PCK@5, mAP@200 drops significantly ($4.82\%$) in case of \textbf{w/o Learnable Weights} (\cref{tab:abal}), depicting its influence on the final feature map quality.

\noindent\textbf{{[iii]} Contribution of $1D$ Convolutions.}
To assess the effectiveness of $1D$ convolution layers, we use simple linear interpolation to match CLIP and SD UNet feature dimensions, before injection. Sharp PCK@5 drop of $25.62\%$ for \textbf{w/o $1D$ Convolutions} (\cref{tab:abal}) reveals its impact on influencing the diffusion denoising process. We posit that the learnable $1D$ convolutions not only match the feature dimension but also \textit{enrich} the CLIP features before injection.

\noindent\textbf{{[iv]} Model variants.}
The performance of our proposed feature extractor depends crucially on the choice of backbone model variant. Accordingly, we explore with multiple pretrained SD and CLIP backbones. \cref{tab:abal} shows SD v2.1 and CLIP ViT-L/14 to achieve the highest PCK@5 and Acc.@1 for sketch-photo correspondence and recognition respectively. Additionally, we also experiment by replacing the CLIP visual encoder of our method with a DINOv2 \cite{oquab2023dinov2} ViT-S/14 encoder for the task of ZS-SBIR, to observe a $32.80\%$ drop in mAP@200. This is probably due to CLIP's \cite{radford2021learning} \textit{multimodal} pertaining (diminishing sketch-photo domain gap), over DINOv2's \cite{oquab2023dinov2} \textit{self-supervised} one.

\noindent\textbf{{[v]} Which timestep value is the most effective?} Different diffusion timesteps yield features with drastically different quality \cite{tang2023emergent, zhang2024tale}. Calculating PCK@5 and mAP@200
\begin{wrapfigure}[8]{l}{0.5\linewidth}
    \centering
    \vspace{-0.45cm}
    \includegraphics[width=1.1\linewidth]{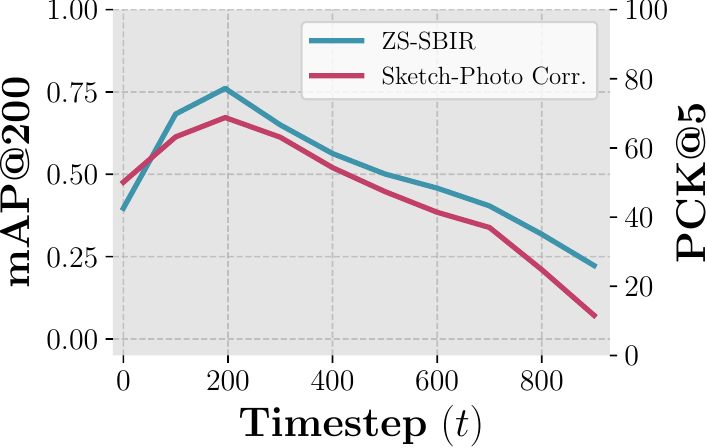}
    \vspace{-0.75cm}
    \caption{Choice of timestep ($t$).}
    \label{fig:timestep}
\end{wrapfigure}
for the tasks of correspondence learning and ZS-SBIR respectively, over varying $t$, we find that $t=195$ produces the best results for both tasks. \cref{fig:timestep} shows our method to be robust to the choice of $t$, where PCK@5 (mAP@200) scores from a wide range of $t$ values consistently surpass the baselines.

\begin{table}[t]
\centering
\renewcommand{\arraystretch}{0.9}

\resizebox{\columnwidth}{!}{
\begin{tabular}{llcc|cc}
\toprule
\multicolumn{2}{c}{\multirow{2}{*}{\textbf{Methods}}} & \textbf{Correspondence} & \textbf{Segmentation} & {\textbf{ZS-SBIR}} & \textbf{Recognition}\\
\cmidrule(lr){3-4}\cmidrule(lr){5-6}
& & \textbf{PCK@5} & \textbf{mIoU} & \textbf{mAP@200} & \textbf{Acc.@1}\\\cmidrule(lr){1-4}\cmidrule(lr){5-6}
\multicolumn{1}{l}{\multirow{3}{*}{SD \cite{rombach2022high}}} & v1.4               & 68.91 & 57.63 & 0.728  & 78.82\\
                                                               & v1.5               & 69.24 & 58.91 & 0.735 & 79.59\\
                                                               & v2.1               & \bf70.31  & \bf60.12 & \bf0.761 & \bf87.02\\
                                                               \cmidrule(lr){1-4}\cmidrule(lr){5-6}
\multicolumn{1}{l}{\multirow{3}{*}{CLIP \cite{radford2021learning}}} & ViT-B/16     & 67.75  & 56.38 & 0.743 & 77.37\\
                                                               & ViT-B/32           & 69.41  & 58.47 & 0.758 & 78.84\\
                                                               & ViT-L/14           & \bf70.31  & \bf60.12 & \bf0.761 & \bf87.02\\
                                                               \cmidrule(lr){1-4}\cmidrule(lr){5-6}
\multicolumn{2}{l}{w/o Aggregation Network}                    & 54.65 & 47.52 & 0.587  & 66.71
\\
\multicolumn{2}{l}{w/o Learnable Weights}                      & 68.75 & 58.13 & 0.726  & 83.79\\
\multicolumn{2}{l}{w/o $1D$ Convolutions}                      & 55.97 & 47.24 & 0.602  & 71.75\\
\rowcolor{MidnightBlue!30}
\multicolumn{2}{l}{\textbf{\textit{Ours-full}}}                & \bf70.31  & \bf60.12 & \bf0.761 & \bf87.02\\
\bottomrule
\end{tabular}
}
\vspace{-0.35cm}
\caption{Ablation on design.}
\label{tab:abal}
\vspace{-0.7cm}
\end{table}

\vspace{-0.2cm}
\section{Conclusion}
\vspace{-0.2cm}
\label{sec:conclusion}
In this paper, we identified SD's limitations in understanding abstract, sparse sketches. Integrating CLIP's semantic features with SD's spatially-aware features, we developed a framework that dynamically combines these strengths. This significantly improves performance across sketch recognition, SBIR, and sketch-based segmentation. Our method highlights the complementary nature of different foundation models and introduces an adaptive feature aggregation network, establishing a new standard for universal sketch feature representation for future research.

{
    \small
    \bibliographystyle{ieeenat_fullname}
    \bibliography{main}
}

\end{document}